\documentclass{article}
\usepackage{ijcai24}

\usepackage{times}
\usepackage{soul}
\usepackage{url}
\usepackage[hidelinks]{hyperref}
\usepackage[utf8]{inputenc}
\usepackage[small]{caption}
\usepackage{graphicx}
\usepackage{amsmath}
\usepackage{amsthm}
\usepackage{booktabs}
\usepackage{algorithm}
\usepackage[switch]{lineno}

\usepackage{xcolor}

\usepackage[noend]{algpseudocode}
\usepackage{kotex}

\usepackage{multirow}
\usepackage{graphicx}
\usepackage{tabularx}
\usepackage{adjustbox}
\usepackage{pifont}

\usepackage{amsmath}
\usepackage{bbm}

\usepackage{array}
\usepackage{mathtools}  
\usepackage{amsfonts}  
\usepackage{comment}
\usepackage{subcaption}
\usepackage{dsfont}

\newcommand{\oursol}{\textsc{GLvSA}}

\newcommand{\SkillEnc}{q_\phi}
\newcommand{\SkillDec}{\pi_\theta^L}
\newcommand{\SkillPrior}{p_\theta}
\newcommand{\SkillD}{\mathcal{P}_\theta} 
\newcommand{\FlatD}{\mathcal{P}_\phi}
\newcommand{\InvD}{\mathcal{P}_\theta^{-1}}
\newcommand{\StateEnc}{E_\theta}
\newcommand{\TargetStateEnc}{E_{\bar{\theta}}}
\newcommand{\StateDec}{D_\phi}

\newcommand{\SubgoalG}{f_\psi}

\newcommand{\SkillPolicy}{\pi_\psi^Z}

\newcommand{\MixedDataset}{\mathcal{D}}
\newcommand{\Traj}{\tau}
\newcommand{\SubTraj}{\tau^\text{sub}}
\newcommand{\SkillPost}{z}

\newcommand{\StateSet}{\mathcal{S}}
\newcommand{\ActionSet}{\mathcal{A}}
\newcommand{\GoalSet}{\mathcal{G}}

\newcommand{\RewardFunc}{R}

\DeclareMathOperator*{\expectation}{\mathbb{E}}
\newcommand{\1}[1]{\mathds{1}{#1}}

\title{Offline Policy Learning via Skill-step Abstraction for Long-horizon Goal-Conditioned Tasks}

\author{
Donghoon Kim$^1$\thanks{Equal contribution.}
\setcounter{footnote}{0}
\and
Minjong Yoo$^2$\footnotemark{}\And
\setcounter{footnote}{1}
Honguk Woo$^{1,2}$\thanks{Corresponding author.}\\
\affiliations
$^1$Department of Artificial Intelligence, Sungkyunkwan University\\
$^2$Department of Computer Science and Engineering, Sungkyunkwan University\\
\emails
\{qwef523, mjyoo2, hwoo\}@skku.com,
}

\begin{document}
\maketitle
\begin{abstract}
Goal-conditioned (GC) policy learning often faces a challenge arising from the sparsity of rewards, when confronting long-horizon goals. 
To address the challenge, we explore skill-based GC policy learning in offline settings, where skills are acquired from existing data and long-horizon goals are decomposed into sequences of near-term goals that align with these skills. 
Specifically, we present an `offline GC policy learning via skill-step abstraction' framework ($\oursol$) tailored for tackling long-horizon GC tasks affected by goal distribution shifts. In the framework, a GC policy is progressively learned offline in conjunction with the incremental modeling of skill-step abstractions on the data. 
We also devise a GC policy hierarchy that not only accelerates GC policy learning within the framework but also allows for parameter-efficient fine-tuning of the policy. 
Through experiments with the maze and Franka kitchen environments, we demonstrate the superiority and efficiency of our $\oursol$ framework in adapting GC policies to a wide range of long-horizon goals. The framework achieves competitive zero-shot and few-shot adaptation performance, outperforming existing GC policy learning and skill-based methods. 
\end{abstract}

\section{Introduction}

In goal-oriented environments, task outcomes are directly related to the achievement of specific goals. This poses a challenge for goal-conditioned (GC) policies when trained through reinforcement learning (RL) algorithms, particularly for tasks with long-horizon goals~\cite{GoFAR,resetfree,actionable}. The challenge is primarily attributed to the reward sparsity associated with achieving long-horizon goals. 
Meanwhile, skill-based RL~\cite{spirl,skimo,jiang2022learning} has exhibited great potential in addressing long-horizon and sparse reward tasks by leveraging skills. These skills are temporally abstracted patterns of expert behaviors spanning over multiple timesteps and they can be acquired offline from expert datasets through imitation learning methods. 

To address the challenge associated with long-horizon goal-reaching tasks, we investigate skill-based RL in the context of offline GC policy learning. Incorporating the learned skills into the procedure of learning a long-horizon GC policy looks promising, as a long-horizon goal can be decomposed into near-term subgoals that can be achievable by the skills. 
However, the nature of imitation-based skill acquisition techniques brings certain issues when they are applied to GC policy learning via RL (GCRL), particularly in scenarios with diverse and varied goals. Imitation-based skills, acquired offline from existing trajectory data, tend to specify a constrained range of achievable goals for GCRL. Thus, this can restrict the adaptation capabilities of GC policies with respect to achievable goals, confining them to the limits of the offline data. This limitation often results in diminished performance in GC tasks, especially those with goal distribution shifts; such shifts occur when there is a discrepancy in the distribution of goals between the dataset used for training and the target goals for evaluation.

In this work, we explore skill-based GC policy learning approaches for long-horizon GC tasks, particularly those affected by goal distribution shifts, by presenting a framework for `offline GC policy learning via skill-step abstraction' ($\oursol$). The $\oursol$ framework harnesses data to obtain skill-based GC policies in offline settings, aiming at enhancing their adaptability and efficiency in the environment with long-horizon GC tasks.  
At its core, the framework employs an iterative process for joint learning of a skill-step model and a GC policy. The model encapsulates expert behaviors in the embeddings of temporally abstracted skills and represents environment dynamics in a skill-aligned latent space. 
This concept of skill-step abstraction in $\oursol$ involves distilling the complexity of long-horizon GC tasks into higher-level, distinct segments (i.e., a set of consecutive action sequences). Each segment or skill-step condenses a series of actions into a more manageable form, facilitating a clearer understanding of the GC task and environment dynamics.  
In each iteration, the model is employed to generate skill-step rollouts, weaving distinct trajectories from the data to derive new imaginary ones. 
These derived trajectories are then used to acquire additional plausible skills, which in turn, broaden the range of achievable goals by GC policies grounded in these skills. This iterative process allows for continual model improvement that addresses the limitation inherently associated with conventional imitation-based skill learning approaches, thereby enhancing the adaptability of skill-based GC policies. 

To accelerate both offline learning and online fine-tuning of GC policies, the framework incorporates a modular structure within the policy hierarchy. This design is centered on generating near-term goals, which are closely aligned with the acquired skills (i.e., skill-step goals). 
The modular structure also enables parameter-efficient policy updates for GC tasks with goal distribution shifts, through separate module optimization for skill-step goal generation. This focused optimization streamlines the adaptation process, making it more effective and efficient (i.e., through few-shot fine-tuning), particularly in environments where goals are constantly evolving.

Through various comparison and ablation experiments with the maze and Franka kitchen environments, we demonstrate that the GC policies learned via our $\oursol$ framework achieve competitive performance for both the zero-shot evaluation and few-shot adaptation scenarios, compared to several state-of-the-art GC policy learning methods (i.e., \cite{GCSL,WGCSL}) and skill-based RL methods (i.e., \cite{spirl,skimo}). 
The contributions of our work are summarized as follows. 
\begin{itemize}
    \item We present $\oursol$, a novel offline policy learning framework using a skill-step model, to improve the adaptability of GC policies upon goal distribution shifts. 
    \item We devise an offline GC policy training scheme, where a skill-step model and a GC policy are jointly learned from data and progressively refined via an iterative process. 
    \item We develop a modular GC policy structure to accelerate both policy learning and fine-tuning. 
    \item We validate the effectiveness of $\oursol$ with zero-shot and few-shot adaptation scenarios, by applying it to navigation and robot manipulation tasks, each presenting distinct challenges in terms of goal distribution shifts. 
    \item Our $\oursol$ stands as the first to incorporate offline skill and model learning in the realm of GCRL.
\end{itemize}

\section{Preliminaries}
\textbf{GCRL} addresses the problem of goal-augmented Markov decision processes (GA-MDPs)  $\mathcal{M}_G$, which are represented by an MDP $\mathcal{M} = \langle \StateSet, \ActionSet, \mathcal{P}, \RewardFunc, \gamma \rangle$ and an additional tuple of a goal space, a goal distribution,  and a state-goal mapping function $\langle \mathcal{G}, p_g, \Phi \rangle$~\cite{GCRL_Survey}.
In a GA-MDP $\mathcal{M}_G$, $\StateSet$ is a state space, $\ActionSet$ is an action space, $\mathcal{P}: \StateSet \times \ActionSet \times \StateSet \rightarrow [0,1]$ is a transition probability, and $\gamma$ is a discount factor. 
A reward function $\RewardFunc:\StateSet \times \ActionSet \times  \mathcal{G}  \rightarrow \mathbb{R}$ is determined by a given goal and is formed as
$\RewardFunc:\StateSet \times \ActionSet \times  \mathcal{G}  \rightarrow \mathbb{R}$ that represents spare rewards; i.e.,

\begin{equation}
    R(s, a, g) = \1(\Phi(s' \sim \mathcal{P}(\cdot| s, a)) = g)
\end{equation}
where $\Phi:\mathcal{S} \rightarrow \GoalSet$ maps states $s$ to a goal $g \in \mathcal{G}$. 
This reward structure can be used for RL-based GC policy learning.
A GC policy such as
$\pi : \StateSet \times \GoalSet \times \ActionSet \rightarrow [0,1]$  
is learned by %
\begin{equation}
\label{eq:obj}
    \expectation_{g \sim p_g} 
    \left[\sum_{t=0}^{T-1} \gamma^t R(s_t, a_t \sim \pi(\cdot | s_t, g), g)\right].
\end{equation} 
In principle, GCRL facilitates multi-task learning by treating distinct tasks as goals, and it also addresses multi-stage complex tasks by incorporating the composite goal of multiple subtasks into its goal representations. 

\textbf{Skill-based RL} is a hierarchical learning paradigm~\cite{HRL} that leverages the notion of skills, which refers to reusable units of prior knowledge and useful behavior patterns extracted from the data. 
This hierarchical RL aims to accelerate the process of policy learning through supervising low-level policies on skills, particularly when tackling long-horizon, sparse reward tasks. In~\cite{spirl}, a skill corresponds to an abstracted $H$-step consecutive action sequence and is represented in a variational auto-encoder (VAE)-based embedding space. Accordingly, skill embeddings $z$ are learned on $H$-step sub-trajectories $\tau^\text{sub}$ such as 
\begin{equation}
\label{eq:subtraj}
    \begin{aligned}
     \tau_{t:t+H} &= (s_t, \dots, s_{t+H}, a_t, \dots, a_{t+H-1}) 
    = (s_{0:H}, a_{0:H-1})
    \end{aligned}
\end{equation}
that are sampled from trajectories $\Traj$ in the data. These embeddings are obtained offline by the \textbf{skill loss} $\mathcal{L}_\text{skill}$ such as
\begin{equation}
\label{eq:vae2}
\begin{aligned}
     \expectation_{\SubTraj \sim \Traj} 
        \Bigg[
            &\sum_{i=0}^{H-1} \underbrace{(\SkillDec\left( s_i, \SkillPost \right) - a_i)^2}_\text{action reconstruction} 
            + \underbrace{\beta \cdot KL(\SkillEnc(z| \tau^{\text{sub}} ) || P(z) )}_\text{regularization}
        \Bigg]
\end{aligned}
\end{equation}
where $KL$ denotes the KL divergence and
$P(z) \sim \mathcal{N}(0,I)$ is a prior distribution of skills.
Here, a skill encoder $\SkillEnc$ converts action sequences to skill embeddings $z$ and a low-level policy $\SkillDec$ serves as a skill decoder, producing an action for a state-skill pair.\footnote{For notation consistency as in Table~\ref{table:notations}, we use the model parameters in subscripts for learnable modules from this point forward. Different subscripts are uniquely meaningful in our approach but do not apply broadly in other related works.}
By this encoding-decoding structure, skills are captured from diverse experiences and establish building blocks that enable an agent to tackle complex tasks effectively. 

Unlike the conventional RL approach that directly selects actions, in skill-based RL, the agent's GC policy $\pi(a | s, g)$ is composed of two modules, a low-level policy $\SkillDec(a | s, z)$ and a skill-based high-level policy $\SkillPolicy(z | s, g)$, i.e., 
\begin{equation}
\label{eq:skillbasedrl}
   \pi(a | s, g) =  \SkillDec(a | s, z)  \circ \SkillPolicy(z | s, g).
\end{equation}
In the hierarchy, the behavior of $\SkillPolicy$ adheres to the constraints of a prior $\SkillPrior$, which regularizes the policy search space, following the entropy maximization RL scheme in~\cite{SAC}. Thus, the learning objective of a GC policy in Eq.~\eqref{eq:obj} can be rewritten using the two policy modules $\SkillDec, \SkillPolicy$ and the prior $\SkillPrior$ optimized by 
\begin{equation}
    \begin{aligned}
    \expectation_{g \sim p_g}  \Bigg[\sum^{T-1}_{t=0} \Bigg[& \sum_{k=tH}^{(t+1)H-1} \left[ R(s_{k}, a_{k} \sim \SkillDec (\cdot | s_{k}, z_{tH} ), g) \right] \\
    &- KL(\SkillPolicy (z_{tH}|s_{tH}, g) || p_\theta(z_{tH}|s_{tH})) \Bigg] \Bigg]
    \end{aligned}
\end{equation}
where $z_{tH} \sim \SkillPolicy (\cdot | s_{tH}, g)$ and $H$ is the skill-step horizon.

\section{Our Framework}
\begin{figure}[t]{
    \centering
    \includegraphics[width=0.48\textwidth]{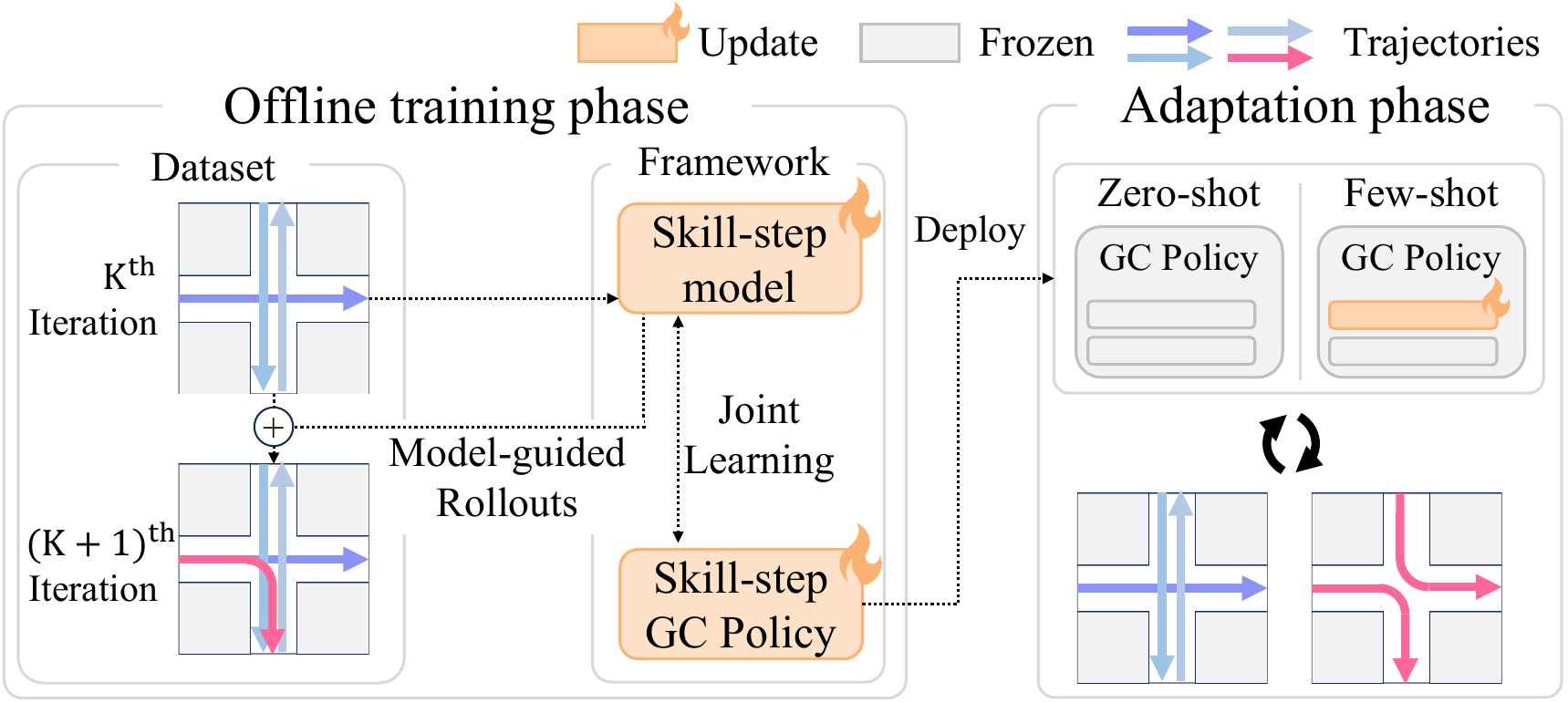}
    \caption{
    GLvSA framework
    }   
    \label{fig:main}
    }
\end{figure}

To address the challenge of tackling long-horizon goal-reaching tasks, we incorporate the skill-based RL strategy into the offline learning process for GC policies. We achieve this integration in our $\oursol$ framework, in which a skill-step model encapsulating skills and environment dynamics at the skill abstraction level, and a GC policy based on the model are jointly and iteratively learned during the \textbf{offline training phase}.
In the online \textbf{adaptation phase}, the GC policy learned offline can be evaluated in a zero-shot manner without any policy updates or it can be parameter-efficiently fine-tuned through few-shot learning. Adaptation is required for the cases when confronting downstream tasks with significant goal distribution shifts.   
Figure~\ref{fig:main} depicts these two phases of $\oursol$; the offline training phase and the adaptation phase.
The offline training phase consists of two tasks using the current dataset: joint learning of the skill-step model and GC policy, and model-guided rollouts. During joint learning, modules for the skill-step model and GC policy 
are updated using trajectories from the dataset. Subsequently, the model-guided rollouts produce trajectories, which are added to the dataset for use in future iterations. 
For downstream GC tasks, the GC policy can be directly deployed and evaluated in a  ``zero-shot'' manner. It can be also fine-tuned via ``few-shot'' updates in a parameter-efficient way to better adapt to a specific task, particularly for the cases subject to goal distribution shifts.
Our GC policy hierarchy employs a modular scheme 
for the policy network. 
This involves dividing the network into three distinct modules, each serving a specific inference function. 
The modular scheme allows for parameter-efficient policy adaptation by confining policy updates within one for skill-step goal generation, thereby maintaining the stability of the other modules. 

\begin{table}[b]
    \centering   
    \begin{tabular}{ccc}
        \toprule
        Type & Modules &  Notation      \\ 
        \midrule
        \multirow{3}{*}{Policy}     &  skill decoder (low-level policy) &  $\SkillDec$ \\
                                    &  skill policy (high-level policy) &  $\SkillPolicy$ \\
                                    &  inverse skill-step dynamics    &   $\InvD$ \\
        \hline
        \multirow{4}{*}{Model}      &  skill-step, flat dynamics  &  $\SkillD$, $\FlatD$  \\
                                    &  skill encoder &  $\SkillEnc$ \\
                                    &  skill prior   &  $\SkillPrior$ \\
                                    &  state encoder, decoder  &  $\StateEnc$, $\StateDec$ \\
    \bottomrule
    \end{tabular}
    \caption{Notations for learnable modules}   
    \label{table:notations}
\end{table}

Table~\ref{table:notations} lists the notations of modules in $\oursol$. They include the skill-based GC \textbf{policy} with low-level policy (skill decoder), high-level policy (skill policy), and inverse skill-step dynamics modules, as well as the skill-step \textbf{model} with dual dynamics, skill encoder, and skill prior modules. These also include a state encoder-decoder, by which the latent state space can align with the skills. 
Each module subscript denotes the role in policy adaptation. $\psi$, $\theta$, and $\phi$ indicate fine-tuned, frozen, and unused modules, respectively, in policy adaptation; e.g., $\SkillEnc$ and $\StateDec$ are used only for the offline training.

\subsection{Skill-step Model Learning}\label{section:SkillStepModel}
\begin{figure}[t]{
    \centering
    \includegraphics[width=0.47\textwidth]{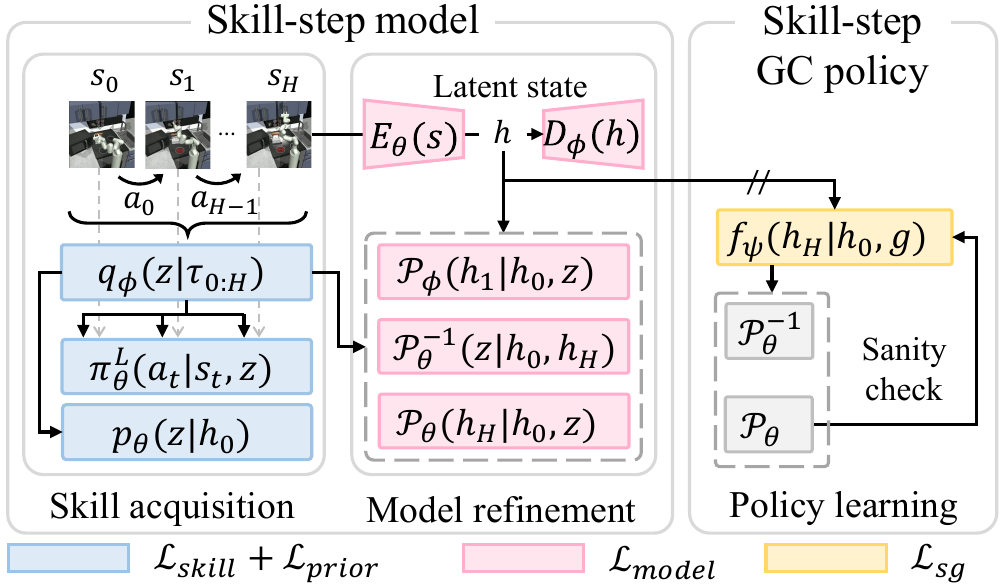}
    \caption{Learning process in GLvSA framework}
    \label{fig:framework}   
}
\end{figure}

To broaden the range of skills and goals beyond the scope of merely imitating individual trajectories in the offline dataset, we employ the iteratively refined skill-step model, involving skill acquisition, model refinement, policy updates, and trajectory generation.
From the iterations onwards, we use the augmented dataset comprising both the original trajectories and those derived from prior iterations' rollouts. In each iteration, all the modules in Table~\ref{table:notations} are jointly optimized, using the losses in Eq.~\eqref{eq:total}, as depicted in Figure~\ref{fig:framework}.
This approach bridges distinct trajectories incrementally in the latent state space, enhancing skill and goal coverage.

\subsubsection{Skill Acquisition}
We use conditional-$\beta$-VAE networks~\cite{cvae,betavae} to obtain skill embeddings $z$, following a similar approach in~\cite{spirl}. A skill encoder $\SkillEnc$ transforms a sub-trajectory $\SubTraj$ in Eq.~\eqref{eq:subtraj} into an embedding $z$ using Eq.~\eqref{eq:vae2}, while a skill decoder $\SkillDec$ reconstructs the sub-trajectory from $z$.
The skill encoder is also updated later through Eq.~\eqref{eq:model}. Furthermore, for minibatches $B = \{\tau^{\text{sub}}_i\}_{i=1}^{N}$ sampled for the dataset $\MixedDataset$, we obtain a skill prior $\SkillPrior$ that infers the distribution of  skills for a given latent state by optimizing the \textbf{skill prior loss}  $\mathcal{L}_\text{prior}$ defined as
\begin{equation}
    \label{eq:prior}
    \begin{aligned}
        \expectation_{B \sim \MixedDataset}
            \left[KL( \SkillPrior (  z| \textbf{h}_t ) \: || \: \textbf{sg}(\SkillEnc(\SkillPost|\SubTraj )) )\right]
    \end{aligned}
\end{equation}
where $\textbf{sg}$ is a stop-gradient function and  $\textbf{h}_t = \textbf{sg}(\StateEnc(s_t))$ is a latent state for the first state $s_t$ in sub-trajectory $\SubTraj$. The skill prior $\SkillPrior$ facilitates rollouts in the latent state space.

\subsubsection{Model Refinement}
To enable rollouts in the latent state space, we jointly optimize state embeddings and two dynamics modules such as skill-step dynamics $\SkillD$ and flat dynamics $\FlatD$, similar to~\cite{skimo}. 
A state encoder $\StateEnc$ represents states in a state embedding space $\mathbb{H}$ that can be reconstructed by a state decoder $\StateDec$. Given a state embedding and a skill, $\FlatD$ predicts the next state embedding by an execution of the single timestep skill, while $\SkillD$ predicts the state by an execution of the $H$-timestep ofskill. 
We also obtain an inverse skill-step dynamics module $\InvD$, which infers a skill for a pair of starting latent state $h_0$ and skill-step latent state $h_H$. These modules (listed in the `Model' row of Table~\ref{table:notations}) are jointly learned by the \textbf{model loss}  $\mathcal{L}_\text{model}$ such as
\begin{equation}
    \label{eq:model}
    \small 
    \begin{aligned}
        &\expectation_{B \sim \MixedDataset}
             \Big[\sum_{k=0}^{H-1} \big[
                    \underbrace{
                        (\StateDec
                            ( h_{t+k}
                            ) - s_{t+k}
                        )^2 }_\text{observation reconstruction} 
                   + \underbrace{
                        ( \FlatD
                            ( h_{t+k}, z
                            ) - \bar{h}_{t+k+1}
                        )^2 }_\text{flat dynamics}\big] \\  
                &+\underbrace{
                    ( \SkillD
                        ( h_t, z
                        ) - \bar{h}_{t+H}
                    )^2}_\text{skill-step dynamics}
                + \underbrace{
                       KL(\textbf{sg}(z) \: || \: \InvD (z| \textbf{h}_t,\textbf{h}_{t+H}))   
                }_\text{inverse skill-step dynamics}\Big] \\
    \end{aligned}    
\end{equation}
where $h_t = E_\theta(s_t)$, $\bar{h}_t = \TargetStateEnc(s_t)$, $\bar{\theta}$ is slowly copied from $\theta$, and $z=q_\phi(\SubTraj)$ is the skill embedding. Note that the skill encoder $\SkillEnc$ is updated by this joint optimization in addition to Eq.~\eqref{eq:skillbasedrl}.
By this, the latent state space $(h \in \mathbb{H})$ becomes closely aligned with the skills, thus making it easier to stitch sub-trajectories of different trajectories.

\subsubsection{Trajectory Generation}
\begin{figure}[t]{
    \centering
    \includegraphics[width=0.47\textwidth]{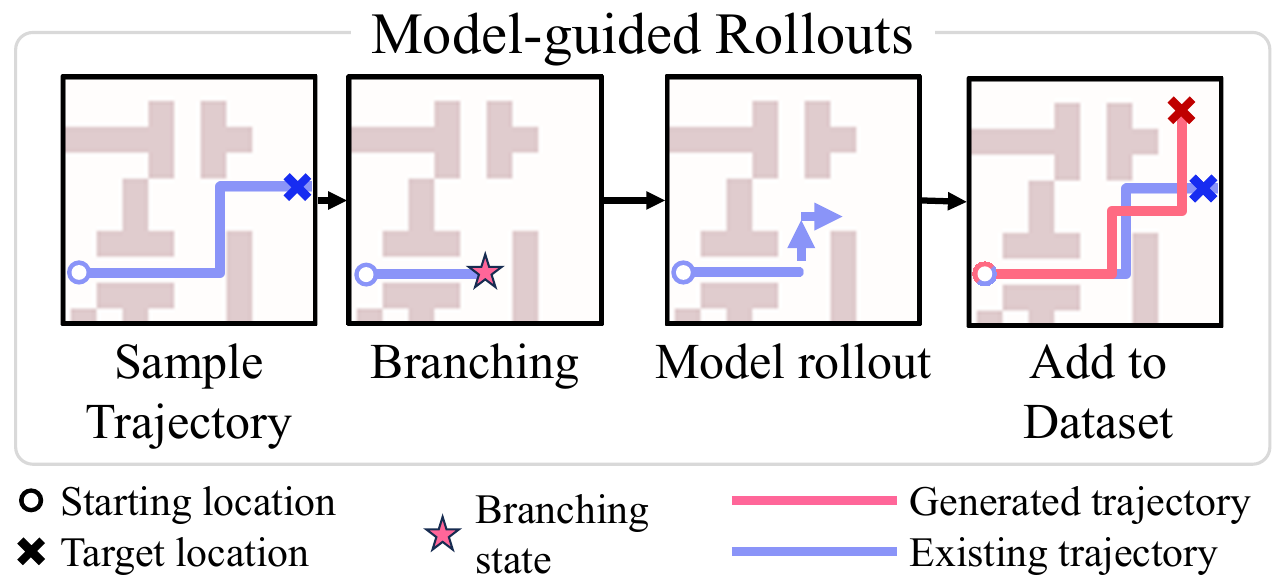}
    \caption{Model-guided rollouts
    }  
    \label{fig:rollout}
}
\end{figure}

Figure~\ref{fig:rollout} illustrates model-guided rollouts. To derive new trajectories in each iteration, we first randomly select branching states from an existing trajectory
(a sequence of state-action pairs). Branching states are used as initial states for the model-guided rollouts. For each branching state, we sample a skill using the skill prior $\SkillPrior$ and then execute the skill to obtain a rollout in the latent space using the flat dynamics $\FlatD$. Subsequently, we transform this rollout outcome, composed of latent variables $h$ and $z$, into an imaginary trajectory of raw state-action pairs $(s, a)$ using the state decoder $\StateDec$ and the low-level policy (skill decoder) $\SkillDec$. These newly derived trajectories are added to the dataset for next iterations and further learning. 
 
\subsection{Skill-step GC Policy Hierarchy}\label{subsec:PolicyLearning}
Following skill-based RL approaches, we structure our GC policy as Eq.~\eqref{eq:skillbasedrl} in a hierarchy with the (high-level) skill policy $\SkillPolicy$ and the (low-level) skill decoder $\SkillDec$.  
To accelerate policy learning and adaptation, we also devise a modular GC policy structure that encompasses a skill-step goal generator $\SubgoalG$ and an inverse skill-step dynamics module $\InvD$. Then, we decompose the skill policy into 
\begin{equation}
    \label{eq:highlevelpolicydef}
    \SkillPolicy(s_t, g) =   \InvD(z | \textbf{h}_t, \hat{h}_{t+H}) \circ \SubgoalG(\hat{h}_{t+H} | \textbf{h}_t, g) \circ \StateEnc(h_t|s_t).
\end{equation}
By this decomposition, for a pair of current state $s_t$ and long-horizon goal $g$, the GC policy first infers a skill-step goal $\hat{h}_{t+H}$ and then obtains its corresponding skill $z$ via the inverse skill-step dynamics.
Note that the skill decoder $\SkillDec$ is learned via reconstruction in Eq.~\eqref{eq:vae2} and it is responsible for translating $z$ into an action sequence. 

In this hierarchy, the inverse skill-step dynamics $\InvD$ is trained as part of the skill-step model in Eq.~\eqref{eq:model}. The adaptation capabilities of the inverse skill-step dynamics rest on such conditions where the environment dynamics remain consistent between the training datasets and downstream tasks; i.e., they share the same underlying world model.
In this respect, for downstream tasks with goal distribution shifts in the same environment, we leverage this hierarchy in a way that only the skill-step goal generator needs to be updated. This facilitates parameter-efficient policy updates in the adaptation phase.   
In this sense,
we optimize the skill-step goal generator $\SubgoalG$ by the \textbf{skill-step goal loss} $\mathcal{L}_{\text{sg}}$ such as
\begin{equation}
    \label{eq:subgoal}
    \begin{aligned}
     \expectation_{B \sim \MixedDataset}
         \bigg[
            \underbrace{ \left( \bar{h}_{t+H} -  f_\psi(\textbf{h}_t, g) \right) ^2}_\text{behavior cloning}+ \underbrace{ \left( f_\psi(\textbf{h}_t, g) - \SkillD (\textbf{h}_t, \hat{z}) \right)^2}_\text{sanity check}         
         \bigg] \\
    \end{aligned}
\end{equation}
where $\bar{h}_{t+H} = \TargetStateEnc(s_{t+H})$ and  $\hat{z} \sim \InvD(\cdot | \textbf{h}_t, \SubgoalG(h_t, g))$.
In Eq.~\eqref{eq:subgoal}, the first term represents behavior cloning error, and the second term serves as a sanity check for cyclic consistency, ensuring that the generated skill-step goal is consistent with the outcome produced by the execution of the skill, which is determined by the skill-step goal. 

\subsection{Offline Training}\label{subsec:offlinetraining}
In each iteration during the training phase, all the learnable modules (9 modules in Table~\ref{table:notations}) in $\oursol$ are jointly optimized by the losses based on skills, a skill prior, environment dynamics, and skill-step goals, incorporating Eq.~\eqref{eq:vae2}-\eqref{eq:subgoal}. 
\begin{equation}
    \label{eq:total}
    \mathcal{L} = \mathcal{L}_\text{skill}\ + \mathcal{L}_\text{prior} + \mathcal{L}_\text{model}  +  \mathcal{L}_\text{sg}
\end{equation}
See Algorithm 1 in Appendix listing the iterative learning procedure for the offline training phase. 

\subsection{Online Adaptation}
After the offline training phase, we can
immediately
construct the GC policy (structured as Eq.~\eqref{eq:skillbasedrl} and~\eqref{eq:highlevelpolicydef}) using the learned modules and evaluate it in a zero-shot manner. 
For downstream tasks with different goal distributions, we can also adapt the policy efficiently through few-shot online updates. In this case, we tune only the skill-step goal generator $\SubgoalG$ via RL, 
while freezing the other modules. The skill-step goal generator $\SubgoalG$ is updated through value prediction-based reward maximization, in addition to prior regularization and state consistency regularization, i.e., 
\begin{equation}
    \label{eq:RL}
    \begin{aligned}    
     \mathbb{E}_{B'} \bigg[
        &\underbrace{-Q(\textbf{h}_t, \SkillPolicy(\textbf{h}_t, g) )}_\text{reward maximization} + \alpha \cdot \underbrace{KL(\SkillPolicy(z | \textbf{h}_t, g) || p_\theta (z|\textbf{h}_t))}_\text{prior regularization}  \\ 
        &+ \underbrace{\left( \SkillD ( h_{t+H} | \textbf{h}_t, \SkillPolicy(\textbf{h}_t, g)) - \SubgoalG(h_{t+H} | \textbf{h}_t, g) \right)^2}_\text{state consistency regularization}
        \bigg]
    \end{aligned}
\end{equation}
where $\textbf{h}_t = \textbf{sg}(\StateEnc(s_t))$ and $B'$ has skill-step transitions $(s_t, z, s_{t+H})$ which are collected online from the environment.
The state consistency regularization ensures the skill-step reachability of the skill-step goal $\SubgoalG(h_{t+H} | \textbf{h}_t, g)$. This regularization is similar to the second term of Eq.~\eqref{eq:subgoal}. For this policy adaptation via RL, the modules parameterized by $\theta$ and $\phi$ except for $\SubgoalG$ remain frozen.  

\section{Experiments}
\subsection{Experiment Settings}
The experiment involves two environments from D4RL~\cite{d4rl}: maze navigation (maze) and  Franka kitchen simulation (kitchen).
In the maze, the agent's objective is to reach a target location from starting location, with states including the agent's coordinates, velocity, and the goal which is defined by pairs of starting and target locations, and rewards are given only upon reaching this target location.
In the Franka kitchen environment, a robot arm manipulates kitchen objects, receiving rewards for achieving the desired state of each target object
such as Microwave, Kettle, Bottom burner, Top burner, Light switch, Slide cabinet, and Hinge cabinet. The robot's state includes its joint and gripper positions, as well as the status of kitchen objects, with goals represented by the states of four target objects.
Both environments use offline datasets for training; i.e., 3,046 trajectories in the maze and 603 in the kitchen from~\cite{Skild} and~\cite{d4rl}, respectively.
In the following experiments, the GC policies' adaptation capabilities are tested under different goal conditions (None, Small, Medium, and Large), which represent the degree of difference between the goal distribution in the offline dataset and that for the adaptation phase.

For evaluation \textbf{metrics}, we adopt a normalized score, ranging from 0 to 100, which is calculated by
\begin{equation}
    \label{eq:norscore}
    \begin{cases} 
        \mathbb{I}\left[ (\Phi(s) - g)^2 \leq 1 \right] \times 100 & \text{for maze}, \\
        (\text{achieved\_obj} / \text{target\_obj})  \times 100 & \text{for kitchen} 
    \end{cases}
\end{equation}
where $\Phi$ maps states $s$ to a specific goal $g$, and achieved\_obj and target\_obj represent the number of target objects successfully manipulated and the number of target objects in the goal, respectively.
The detailed settings for goal specifications and goal distribution shifts are in Appendix A. 

\begin{table*}[t!]
    \centering
    \begin{tabular}{@{\extracolsep{2pt}}ccccccc}
    \toprule
    Environment                      & Dist. shift    &  GCSL         & WGCSL          & SPiRL+GCSL     & SkiMo+GCSL       &  $\oursol$ (Ours)      \\ 
    \midrule
    \multirow{4}{*}{Maze}    & None       &$73.7\pm 2.0$   &$46.2\pm4.1$    &$23.4 \pm 6.4$  &$98.8 \pm 1.1$    &$\textbf{100.0}\pm\textbf{0.0}$   \\
                             & Small          &$33.3\pm 4.0$   &$15.2\pm10.9$   &$26.9 \pm 3.0$  &$74.9 \pm 1.1$    & $\textbf{89.5}\pm\textbf{2.0}$   \\
                             & Medium         &$12.6\pm 5.2$   &$14.4\pm3.6$    &$23.2 \pm 5.2$  &$74.0 \pm 2.5$    &$\textbf{84.2}\pm\textbf{7.3}$   \\
                             & Large          &$5.3\pm 3.9$   &$1.3\pm 1.3$     &$15.8 \pm 3.9$  &$32.5 \pm 8.7$    &$\textbf{66.7}\pm \textbf{7.0}$    \\
    \midrule
                             
    \multirow{4}{*}{Kitchen} & None           &$35.9\pm4.8$    &$69.6\pm 1.1$   &$49.1 \pm 2.5$  &$66.6 \pm 7.5$   &$\textbf{98.7}\pm \textbf{1.3}$  \\
                             & Small          &$0.0\pm0.0$     &$44.7\pm 1.8$   &$26.2 \pm 8.5$  &$54.8 \pm 5.7$   &$\textbf{88.6}\pm\textbf{1.3}$  \\   
                             & Medium         &$0.2\pm0.2$     &$31.4\pm 0.8$   &$31.2 \pm 1.7$  &$47.7 \pm 4.3$   &$\textbf{86.8}\pm\textbf{1.7}$  \\
                             & Large          &$0.4\pm0.2$     &$18.3\pm 1.0$   &$28.7 \pm 7.0$  &$41.7 \pm 5.2$   &$\textbf{72.2}\pm\textbf{3.4}$ \\
    \bottomrule
    \end{tabular}
    \caption{Zero-shot evaluation performance}    
    \label{table:zeroshot}
\end{table*}

%
For comparison, we use several \textbf{baselines}, encompassing state-of-the-art GC policy learning and skill-based RL methods. 
Our GC problem settings are characterized by two main constraints: (1) offline GC policy learning, wherein data collection by the policy is prohibited during the training phase, and (2) online GC policy adaptation, which permits either zero-shot evaluation or few-shot updates. The policy adaptation aims at tackling a range of tasks with different degrees of goal distribution shifts.
Given the aforementioned constraints, we adapt several baselines, which typically rely on online environment interactions and data collection, to function with offline pre-trained GC policies. 

\textbf{GCSL}~\cite{GCSL} is used as a baseline for GCRL experiments to evaluate offline GC policy learning without skill-based strategies. It iteratively collects data through online environment interactions and performs hindsight relabeling and imitation learning to optimize the GC policy. In the experiments, online data collection is replaced with an expert dataset for offline settings, similar to~\cite{WGCSL,GoFAR}.
\textbf{WGCSL}~\cite{WGCSL} is a GCSL variant, employing the advantage-based behavior cloning weights.
This method is used to evaluate the state-of-the-art performance of offline GCSL.
\textbf{SPiRL+GCSL} is a GC variant of the skill-based RL method, SPiRL~\cite{spirl} to evaluate the GC performance of skill-based RL in offline settings.      
In our experiments, we train a GC policy using the skills in a supervised manner. We also incorporate the GCSL's objective into the skill-based RL objective structure. 
\textbf{SkiMo+GCSL} is a GC variant of the skill-based model-based RL, SkiMo~\cite{skimo} which jointly optimizes a model and skills. We adapt SkiMo with GC policy training, and combine the objectives of GCSL and skill-based RL, in the same way of SPiRL+GCSL. 
This method relies solely on the offline data for model and skill learning, unlike our approach which employs iterative learning through the model-guided skill rollouts.  
It shows the state-of-the-art GC performance of skill-based RL in offline settings.

\subsection{Zero-shot Evaluation Performance}
\begin{figure}[t]{
    \centering
    \includegraphics[width=0.45\textwidth]{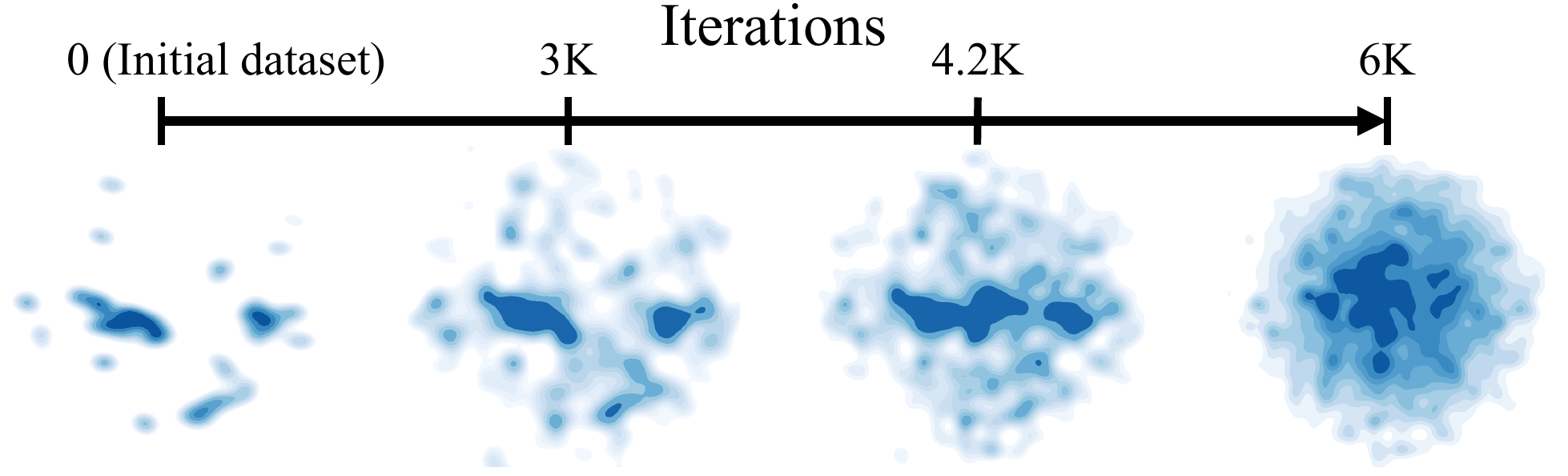}
    \caption{Goal coverage expansion (visualized in T-SNE)}   
    \label{fig:goal_coverage}
    }
\end{figure}

In zero-shot scenarios, we evaluate the GC policy of each method without any policy updates. 
Table~\ref{table:zeroshot} reveals that our $\oursol$ achieves superior zero-shot performance in both maze and kitchen environments, where the ``Dist. shift'' column specifies different degrees of goal distribution shifts between the offline dataset and the evaluation tasks. 
%
Compared with the most competitive baseline SkiMo+GCSL, $\oursol$  demonstrates a gain of $14.6$, $20.2$, and $34.2$ in normalized scores for Small, Medium, and Large distribution shifts, respectively, in the maze environment, as well as a gain of $33.8$, $39.1$, $30.5$, respectively, in the kitchen. 
As the degree of distribution shifts increases, we observe a greater performance drop, particularly for the action-step baselines (GCSL, WGCSL) which do not incorporate skill-based strategies. In contrast, $\oursol$ displays robustness against these distribution shifts, demonstrating the ability of our GC policy to generalize across a wider range of goals. 
Importantly, our approach employs the skill-step model to guide trajectory generation, using iterative joint learning of skills, goals, and the model. This renders the skills and goals obtained beyond the scope of the initial offline dataset, marking a distinction from the other skill-based baselines such as SPiRL+GCSL and SkiMo+GCSL, which more rely on the dataset. 
Figure~\ref{fig:goal_coverage} illustrates the expanding goal coverage in the latent state space of the kitchen environment, achieved by $\oursol$ over iterations. This growth indicates that a variety of goals are being learned through trajectory generation, forming the basis for strong zero-shot performance.

\begin{table}[h]
    \centering
    \begin{adjustbox}{width=0.45 \textwidth}
    \begin{tabular}{@{\extracolsep{2pt}}cccccc}
    \toprule
    Env.                 & Shot       & SPiRL+GCSL        & SkiMo+GCSL     &  $\oursol$       \\ 
    \midrule
    \multirow{4}{*}{Maze}   &  1  & $7.5 \pm 5.2$  & $16.2 \pm 12.0$  & $\textbf{67.1} \pm \textbf{14.4}$    \\
                            &  10  & $3.9 \pm 0.0$  & $15.8 \pm 6.5$  & $\textbf{74.1} \pm \textbf{5.7}$    \\
                            &  25  & $15.8 \pm 15.5$  & $40.8 \pm 15.5$  & $\textbf{73.7} \pm \textbf{5.7}$    \\
                            &  50  & $46.1 \pm 27.8$  & $61.0 \pm 11.9$  & $\textbf{76.3} \pm \textbf{5.5}$    \\
    \midrule
    \multirow{4}{*}{Kitchen} &  1  &$ 23.4 \pm 1.7 $   &$ 18.6 \pm 5.4 $  & $\textbf{68.0} \pm \textbf{5.9}$ \\
                             &  10 & $24.4 \pm 2.9 $   &$ 20.3 \pm 2.5 $  & $\textbf{78.4} \pm \textbf{4.2}$ \\
                            &  25  & $22.8 \pm 3.1$  & $34.8\pm 1.0$   & $\textbf{83.5} \pm \textbf{8.8}$    \\
                            &  50  & $25.1 \pm 2.8$  & $54.2 \pm 4.2$   & $\textbf{92.5} \pm \textbf{2.3}$    \\
    \bottomrule              
    \end{tabular}
    \end{adjustbox}
    \caption{Few-shot adaptation performance. This includes the competitive baselines; the other results are in Appendix.}    
    \label{table:fewshot}
\end{table}

\subsection{Few-shot Adaptation Performance}
Table~\ref{table:fewshot} presents the adaptation performance in normalized scores, obtained through few-shot policy updates, with each shot representing an episode (e.g., in the ``Shot'' column, ``Shot = 1'' specifies that the policy is updated via the environment interaction of a single episode).
In these few-shot adaptation scenarios, we focus on the cases with significant distribution shifts (i.e., ``Dist. shift = Large''). Such shifts tend to diminish the zero-shot performance of GC policies, as observed in Table~\ref{table:zeroshot} previously. 
As shown,  $\oursol$ consistently outperforms the baselines across the shots.
For smaller shot counts (1, 10), $\oursol$ outperforms the most competitive baseline (SkiMo+GCSL) by margins of $50.9$ and $58.3$ in the maze, and $49.4$ and $58.1$ in the kitchen, respectively. 
While the baseline's 1-shot and 10-shots performance degrades, compared to its zero-shot performance, $\oursol$ exhibits comparable performance by 1-shot and an improvement by 10-shots, compared to its respective zero-shot; e.g., in the kitchen environment with Large shift, our performance increases from a zero-shot score of 72.2 to 78.4 at 10-shots. 
As the shot counts increase to 25 and 50, the performance gaps between our $\oursol$ and the baseline narrow to $32.9$ and $15.3$ in the maze, and $48.7$ and $38.3$ in the kitchen, respectively. Nevertheless, these gaps remain substantial. Such results highlight the efficiency of our GC policy hierarchy, which facilitates parameter-efficient updates specifically targeting the skill-step goal generation, thereby enabling few-shot policy adaptation. 

\subsection{Ablations}
We conduct ablation studies in the kitchen environment. 

{\textbf{Skill-step model structure.}}
Table~\ref{table:skillstep_model} contrasts the zero-shot performance by our model structure employing the skill-step dynamics $\SkillD$ for model-guided rollouts against a conventional model structure that solely relies on the flat dynamics $\FlatD$. We also include the ``No rollout'' case.
This comparison sheds light on the advantages of using the skill-step dynamics. Our approach (w/ $\SkillD$) consistently yields gains over its counterpart (w/o $\SkillD$). For the cases of larger distribution shifts, we observe a significant performance drop of (w/o $\SkillD$), which is attributed to the compounding error, a known drawback of model-based RL approaches~\cite{janner2019trust,resetfree}.
The integration of $\SkillD$ into the model regulates the latent state space using the skills, enabling more dependable rollouts. This, in turn, facilitates enhancements in skill acquisition, model refinement, and policy optimization, even upon substantial goal distribution shifts. We observe that (w/o $\SkillD$) performs worse than the No rollout case, which is also associated with the compounding error effect. 
The impact of the model errors can be more pronounced than the benefits of enhanced diversity from trajectory generation.

\begin{table}[h]
    \centering   
    {    
    \begin{tabular}{@{\extracolsep{2pt}}cccc}
        \toprule
        Dist. shift             & No rollout         & (w/o $\SkillD$)   & $\oursol$ (w/ $\SkillD$)          \\ 
        \midrule
        \multirow{1}{*}{Small}  &  $78.4 \pm 2.0$   & $77.6 \pm 4.2$  & $\textbf{88.6} \pm \textbf{1.3}$  \\
        \multirow{1}{*}{Medium} &  $64.7 \pm 1.5$   & $48.9 \pm 3.3$  & $ \textbf{86.8} \pm \textbf{1.7}$    \\
        \multirow{1}{*}{Large}  &  $57.7 \pm 2.9$   & $51.9 \pm 6.7$  & $\textbf{72.2} \pm \textbf{3.4}$   \\
        \bottomrule
    \end{tabular}
    }
    \caption{Effect of skill-step dynamics $\SkillD$}
    \label{table:skillstep_model}
\end{table}

{\textbf{Skill horizon.}
Table~\ref{table:skill_length} specifies the impact of skill horizon $H$ (defined in Eq.~\eqref{eq:subtraj}) on the zero-shot performance, where $H$ sets to 1, 5, 10, and 40 in timesteps. We observe that the performance peaks at $H = 10$.
As the skill horizon $H$ decreases, the temporal abstraction capability of skills diminishes, leading to a performance drop.
This is because the range regularized by the skill-step dynamics $\SkillD$ in the latent state space contracts, which in turn impacts the state and skill alignment. 
On the contrary, the larger horizon $H = 40$  incurs the difficulty for learning the skill-step dynamics $\SkillD$ and goal generator $\SubgoalG$, thus resulting in a performance decline.

\begin{table}[h]
    \centering   
    \begin{adjustbox}{width=0.48 \textwidth}{   
    \begin{tabular}{@{\extracolsep{2pt}}ccccc}
        \toprule
        Dist. shift   & H =1            &   H=5           & H=10 (Ours)        & H=40          \\ 
        \midrule
        None  & $44.8\pm 1.4$   & $92.8 \pm 1.9$  & $\textbf{98.7} \pm \textbf{1.3}$ & $90.1 \pm 3.0$  \\
        Small         & $41.4\pm 1.5$   & $87.4 \pm 1.0$   & $\textbf{88.6} \pm \textbf{1.3}$ & $76.9 \pm 1.1$  \\
        Medium        & $32.1\pm 3.1$   & $78.8 \pm 1.1$   & $\textbf{86.8} \pm \textbf{1.7}$ & $65.8 \pm 2.0$  \\
        Large         & $29.5\pm 1.6$   & $66.7 \pm 0.5$   & $\textbf{72.2} \pm \textbf{3.4}$ & $60.2 \pm 2.5$  \\
        \bottomrule
    \end{tabular}
    }
    \end{adjustbox}
    \caption{Effect of skill horizon $H$}
    \label{table:skill_length}
\end{table}

{\textbf{Skill-step goal generation.}
To assess the effectiveness of the policy hierarchy incorporating the skill-step goal generation, we compare the adaptation performance of the proposed policy hierarchy in Section~\ref{subsec:PolicyLearning} to its counterpart implemented without the skill-step goal generation (w/o $\SubgoalG$). In addition, we introduce another variant, denoted as (w/ 2-skill-steps), which specifically employs a goal state two skill-steps away. This contrasts with our approach that not only uses a goal state a single skill-step away (a skill-step goal) but also harnesses the alignment of skills and goals in terms of temporal abstraction. 
Table~\ref{table:policy_hierarchy} shows that our policy hierarchy outperforms both the counterpart and the variant.
Our policy hierarchy effectively mitigates the exploration inefficiency that emerges due to multiple skill combinations to achieve a long-horizon goal, exploiting the skill-step goal generation.

\begin{table}[h]
    \centering   
    \begin{adjustbox}{width=0.45\textwidth}   
    {    
    \begin{tabular}{@{\extracolsep{2pt}}cccc}
        \toprule
        Shot & (w/o $\SubgoalG$) &  (w/ 2-skill-steps) & $\oursol$ (w/ $\SubgoalG$) \\
        \midrule
        1     & $52.3 \pm 8.3$ & $57.3 \pm 7.6$  & $\textbf{68.0} \pm \textbf{5.9}$ \\
        10    & $56.0 \pm 10.9$ & $62.1 \pm 8.9$  & $\textbf{78.4} \pm \textbf{4.2}$ \\      
        25    & $58.9 \pm 5.0$ & $60.2 \pm 5.9$  & $\textbf{83.5} \pm \textbf{8.8}$ \\
        50    & $63.9 \pm 17.1$ & $70.8 \pm 4.6$  & $\textbf{92.5} \pm \textbf{2.3}$ \\
        \bottomrule
    \end{tabular}
    }
    \end{adjustbox}
    \caption{Effect of skill-step goal generation}
    \label{table:policy_hierarchy}
\end{table}

{\textbf{Sanity check by cyclic consistency.}
Table~\ref{table:sanity_check} shows the effect of the sanity check in Eq.~\eqref{eq:subgoal} which ensures the cyclic consistency between the deduced skill-step goals and the outcomes of skill executions. 
We compare an implementation using only the behavior cloning term without the sanity check (denoted as BC) to our approach (w/ sanity check). We also implement another method (denoted as BC+SR) in which the high-level policy $\SkillPolicy$ is regularized by skill embeddings $z$ through KL-Divergence with the inverse skill-step dynamics, similar to SAC~\cite{SAC} and SPiRL~\cite{spirl}.
Our approach outperforms both BC and BC+SR across all degrees of distribution shifts.
The sanity check regularizes $\SkillPolicy$ robustly, assisting the skill-step goal generator $\SubgoalG$ to predict achievable near-term goals by the skills.

\begin{table}[t]
    \centering   
    {    
    \begin{tabular}{@{\extracolsep{2pt}}cccc}
        \toprule
        Dist. shift   & BC                                       & BC+SR                    & $\oursol$            \\ 
        \midrule
        None         & $95.6 \pm 0.8$         & $85.5 \pm 2.4$           & $\textbf{98.7} \pm \textbf{1.3}$                    \\

        Small         & $87.7 \pm 0.5$   & $75.4 \pm 2.3$    & $\textbf{88.6} \pm \textbf{1.3}$  \\
        Medium        & $75.7 \pm 3.2$   & $58.2 \pm 4.6$    & $\textbf{86.8} \pm \textbf{1.7}$                    \\
        Large         & $67.3 \pm 4.0 $  & $50.0 \pm 0.5 $    & $\textbf{72.2} \pm \textbf{3.4}$  \\
        \bottomrule
    \end{tabular}
    }
    \caption{Effect of sanity check}
    \label{table:sanity_check}
\end{table}

\section{Related Work}
\textbf{GCRL} represents one of the RL approaches for solving diverse tasks using reward-relevant goal representations~\cite{GCRL_Survey,wang2023d}.
Several GCRL methodologies, particularly for addressing long-horizon goals that inherently incur the reward sparsity, have been introduced, including graph-based planning~\cite{kim2023imitating,MoCoDa}, incremental goal generation~\cite{cho2023outcomedirected}, and subgoal generation~\cite{PTP}.
Our skill-step model approach also tackles the long-horizon goal problem, exploring the skill-based RL strategy in GCRL contexts. 

\textbf{Skill-based RL} exploits the skill-level abstraction of expert action sequences to facilitate the online process of policy learning, leveraging task-agnostic trajectory data in offline skill learning~\cite{spirl,nair2020awac,hussonnois2023controlled,jiang2022learning}. This skill-based strategy has been applied in meta-learning~\cite{simpl}, model-based learning~\cite{skimo}, and cross-domain settings~\cite{star}.
Yet, it has not been fully explored to adopt the skill-based strategy in GCRL. Particularly, upon goal distribution shifts of target tasks, both the learned skills and policies using the skills are required to be retrained. 
We employ the model-guided goal exploration at the skill-level, hence enhancing the generalization of a learned GC policy against goal distribution shifts.

In \textbf{model-based RL} and planning~\cite{dreamer,skimo,egorov2022scalable,hafner2020mastering}, several methods have been introduced for facilitating imaginary exploration~\cite{wang2021offline,mendonca2021discovering,hu2023planning}. 
In GCRL contexts, the model predictive control employs subgoal generation~\cite{TDMPC}, and the skill embedding is integrated into long-horizon environments~\cite{skimo}. Unlike these methods directly exploiting the model for goal learning either at the action-level or the skill-level, we devise effective model-based goal learning that uses the cyclic consistency of skills and near-term goals, thus broadening the goal range and enabling to generalize the GC policy against goal distribution shifts.

\section{Conclusion}
In this work, we presented a novel offline GC policy learning framework designed to address long-horizon GC tasks subject to goal distribution shifts. By employing the skill-step model-guided rollouts, our framework extends the range of achievable goals and enhances the adaptation capabilities of the GC policy within offline scenarios. We also proposed a modular policy hierarchy that features the ability to generate near-term goals, attainable by the skills. This hierarchy bolsters both the robustness of the GC policy and its adaptation efficiency against goal distribution shifts.
In our future work, we aim to extend the skill-step model for more complex circumstances where distribution shifts in both environment dynamics and goals occur concurrently. Our focus will be on generalizing the skills to better capture the intricacies of environment dynamics patterns during the training phase. 

\appendix

\section*{Acknowledgements}
We would like to thank anonymous reviewers for their valuable comments and suggestions. 
This work was supported by 
Institute of Information \& communications Technology Planning \& Evaluation (IITP) grant funded by the Korea government (MSIT) (No.
2022-0-01045, 
2022-0-00043,
2020-0-01821,
2019-0-00421) 
 and by the National Research Foundation of Korea (NRF) grant funded by the MSIT 
(No. RS-2023-00213118, NRF-2020M3C1C2A01080819) 
and by Samsung electronics.

\section*{Contribution Statement}
Donghoon Kim and Minjong Yoo made equal contributions.
All the authors participated in designing research, performing
research, analyzing data, and writing the paper.

\bibliographystyle{named}
\bibliography{main}

\begin{thebibliography}{}

\bibitem[\protect\citeauthoryear{Chebotar \bgroup \em et al.\egroup }{2021}]{actionable}
Yevgen Chebotar, Karol Hausman, Yao Lu, Ted Xiao, Dmitry Kalashnikov, Jacob Varley, Alex Irpan, Benjamin Eysenbach, Ryan~C Julian, Chelsea Finn, et~al.
\newblock Actionable models: Unsupervised offline reinforcement learning of robotic skills.
\newblock In {\em Proc.of the 38th International Conference on Machine Learning (ICML 2021)}, pages 1518--1528. PMLR, 2021.

\bibitem[\protect\citeauthoryear{Cho \bgroup \em et al.\egroup }{2023}]{cho2023outcomedirected}
Daesol Cho, Seungjae Lee, and H.~Jin Kim.
\newblock Outcome-directed reinforcement learning by uncertainty {\textbackslash}\& temporal distance-aware curriculum goal generation.
\newblock In {\em Proc.of the 11th International Conference on Learning Representations (ICLR 2023)}, 2023.

\bibitem[\protect\citeauthoryear{Egorov and Shpilman}{2022}]{egorov2022scalable}
Vladimir Egorov and Alexei Shpilman.
\newblock Scalable multi-agent model-based reinforcement learning.
\newblock In {\em Proc.of the 21st International Conference on Autonomous Agents and Multiagent Systems (AAMAS 2022)}, pages 381--390, 2022.

\bibitem[\protect\citeauthoryear{Fang \bgroup \em et al.\egroup }{2022}]{PTP}
Kuan Fang, Patrick Yin, Ashvin Nair, and Sergey Levine.
\newblock Planning to practice: Efficient online fine-tuning by composing goals in latent space.
\newblock In {\em Proc.of the 35th IEEE/RSJ International Conference on Intelligent Robots and Systems (IROS 2022)}, pages 4076--4083. IEEE, 2022.

\bibitem[\protect\citeauthoryear{Fu \bgroup \em et al.\egroup }{2020}]{d4rl}
Justin Fu, Aviral Kumar, Ofir Nachum, George Tucker, and Sergey Levine.
\newblock D4rl: Datasets for deep data-driven reinforcement learning.
\newblock {\em arXiv preprint arXiv:2004.07219}, 2020.

\bibitem[\protect\citeauthoryear{Ghosh \bgroup \em et al.\egroup }{2021}]{GCSL}
Dibya Ghosh, Abhishek Gupta, Ashwin Reddy, Justin Fu, Coline~Manon Devin, Benjamin Eysenbach, and Sergey Levine.
\newblock Learning to reach goals via iterated supervised learning.
\newblock In {\em Proc of the 9th International Conference on Learning Representations (ICLR 2021)}, 2021.

\bibitem[\protect\citeauthoryear{Haarnoja \bgroup \em et al.\egroup }{2018}]{SAC}
Tuomas Haarnoja, Aurick Zhou, Pieter Abbeel, and Sergey Levine.
\newblock Soft actor-critic: Off-policy maximum entropy deep reinforcement learning with a stochastic actor.
\newblock In {\em Proc.of the 35th International conference on machine learning (ICML 2018)}, pages 1861--1870. PMLR, 2018.

\bibitem[\protect\citeauthoryear{Hafner \bgroup \em et al.\egroup }{2019}]{dreamer}
Danijar Hafner, Timothy Lillicrap, Jimmy Ba, and Mohammad Norouzi.
\newblock Dream to control: Learning behaviors by latent imagination.
\newblock {\em arXiv preprint arXiv:1912.01603}, 2019.

\bibitem[\protect\citeauthoryear{Hafner \bgroup \em et al.\egroup }{2020}]{hafner2020mastering}
Danijar Hafner, Timothy~P Lillicrap, Mohammad Norouzi, and Jimmy Ba.
\newblock Mastering atari with discrete world models.
\newblock In {\em Proc.of the 8th International Conference on Learning Representations (ICLR 2020)}, 2020.

\bibitem[\protect\citeauthoryear{Hansen \bgroup \em et al.\egroup }{2022}]{TDMPC}
Nicklas~A Hansen, Hao Su, and Xiaolong Wang.
\newblock Temporal difference learning for model predictive control.
\newblock In {\em Proc.of the 39th International Conference on Machine Learning (ICML 2022)}, pages 8387--8406. PMLR, 2022.

\bibitem[\protect\citeauthoryear{Higgins \bgroup \em et al.\egroup }{2017}]{betavae}
Irina Higgins, Loic Matthey, Arka Pal, Christopher Burgess, Xavier Glorot, Matthew Botvinick, Shakir Mohamed, and Alexander Lerchner.
\newblock beta-vae: Learning basic visual concepts with a constrained variational framework.
\newblock In {\em Proc.of the 5th International conference on learning representations (ICLR 2017)}, 2017.

\bibitem[\protect\citeauthoryear{Hu \bgroup \em et al.\egroup }{2023}]{hu2023planning}
Edward~S. Hu, Richard Chang, Oleh Rybkin, and Dinesh Jayaraman.
\newblock Planning goals for exploration.
\newblock In {\em Proc.of the 11th International Conference on Learning Representations (ICLR 2023)}, 2023.

\bibitem[\protect\citeauthoryear{Hussonnois \bgroup \em et al.\egroup }{2023}]{hussonnois2023controlled}
Maxence Hussonnois, Thommen~George Karimpanal, and Santu Rana.
\newblock Controlled diversity with preference: Towards learning a diverse set of desired skills.
\newblock In {\em Proc.of the 22nd International Conference on Autonomous Agents and Multiagent Systems (AAMAS 2023)}, pages 1135--1143, 2023.

\bibitem[\protect\citeauthoryear{Janner \bgroup \em et al.\egroup }{2019}]{janner2019trust}
Michael Janner, Justin Fu, Marvin Zhang, and Sergey Levine.
\newblock When to trust your model: Model-based policy optimization.
\newblock In {\em Proc.of the 33rd Advances in neural information processing systems (NeurIPS 2019)}, volume~32, 2019.

\bibitem[\protect\citeauthoryear{Jiang \bgroup \em et al.\egroup }{2022}]{jiang2022learning}
Yiding Jiang, Evan Liu, Benjamin Eysenbach, J~Zico Kolter, and Chelsea Finn.
\newblock Learning options via compression.
\newblock In {\em Proc.of the 36th Advances in Neural Information Processing Systems (NeurIPS 2022)}, volume~35, pages 21184--21199, 2022.

\bibitem[\protect\citeauthoryear{Kim \bgroup \em et al.\egroup }{2023}]{kim2023imitating}
Junsu Kim, Younggyo Seo, Sungsoo Ahn, Kyunghwan Son, and Jinwoo Shin.
\newblock Imitating graph-based planning with goal-conditioned policies.
\newblock In {\em Proc.of the 11th International Conference on Learning Representations (ICLR 2023)}, 2023.

\bibitem[\protect\citeauthoryear{Liu \bgroup \em et al.\egroup }{2022}]{GCRL_Survey}
Minghuan Liu, Menghui Zhu, and Weinan Zhang.
\newblock Goal-conditioned reinforcement learning: Problems and solutions.
\newblock {\em arXiv preprint arXiv:2201.08299}, 2022.

\bibitem[\protect\citeauthoryear{Lu \bgroup \em et al.\egroup }{2020}]{resetfree}
Kevin Lu, Aditya Grover, Pieter Abbeel, and Igor Mordatch.
\newblock Reset-free lifelong learning with skill-space planning.
\newblock In {\em Proc.of the 8th International Conference on Learning Representations (ICLR 2020)}, 2020.

\bibitem[\protect\citeauthoryear{Ma \bgroup \em et al.\egroup }{2022}]{GoFAR}
Jason~Yecheng Ma, Jason Yan, Dinesh Jayaraman, and Osbert Bastani.
\newblock Offline goal-conditioned reinforcement learning via $ f $-advantage regression.
\newblock In {\em Proc.of the 36th Advances in Neural Information Processing Systems (NeurIPS 2022)}, volume~35, pages 310--323, 2022.

\bibitem[\protect\citeauthoryear{Mendonca \bgroup \em et al.\egroup }{2021}]{mendonca2021discovering}
Russell Mendonca, Oleh Rybkin, Kostas Daniilidis, Danijar Hafner, and Deepak Pathak.
\newblock Discovering and achieving goals via world models.
\newblock In {\em Proc.of the 35th Advances in Neural Information Processing Systems (NeurIPS 2021)}, volume~34, pages 24379--24391, 2021.

\bibitem[\protect\citeauthoryear{Nair \bgroup \em et al.\egroup }{2020}]{nair2020awac}
Ashvin Nair, Abhishek Gupta, Murtaza Dalal, and Sergey Levine.
\newblock Awac: Accelerating online reinforcement learning with offline datasets.
\newblock {\em arXiv preprint arXiv:2006.09359}, 2020.

\bibitem[\protect\citeauthoryear{Nam \bgroup \em et al.\egroup }{2022}]{simpl}
Taewook Nam, Shao-Hua Sun, Karl Pertsch, Sung~Ju Hwang, and Joseph~Jaewhan Lim.
\newblock Skill-based meta-reinforcement learning.
\newblock In {\em Proc.of the 10th International Conference on Learning Representations (ICLR 2022)}, 2022.

\bibitem[\protect\citeauthoryear{Pertsch \bgroup \em et al.\egroup }{2021}]{spirl}
Karl Pertsch, Youngwoon Lee, and Joseph Lim.
\newblock Accelerating reinforcement learning with learned skill priors.
\newblock In {\em Proc.of the 5th Conference on robot learning (CoRL 2021)}, pages 188--204. PMLR, 2021.

\bibitem[\protect\citeauthoryear{Pertsch \bgroup \em et al.\egroup }{2022}]{Skild}
Karl Pertsch, Youngwoon Lee, Yue Wu, and Joseph~J Lim.
\newblock Guided reinforcement learning with learned skills.
\newblock In {\em Proc.of the 6th Conference on Robot Learning (CoRL 2022)}, pages 729--739. PMLR, 2022.

\bibitem[\protect\citeauthoryear{Pertsch \bgroup \em et al.\egroup }{2023}]{star}
Karl Pertsch, Ruta Desai, Vikash Kumar, Franziska Meier, Joseph~J Lim, Dhruv Batra, and Akshara Rai.
\newblock Cross-domain transfer via semantic skill imitation.
\newblock In {\em Proc.of the 7th Conference on Robot Learning (CoRL 2023)}, pages 690--700. PMLR, 2023.

\bibitem[\protect\citeauthoryear{Pitis \bgroup \em et al.\egroup }{2022}]{MoCoDa}
Silviu Pitis, Elliot Creager, Ajay Mandlekar, and Animesh Garg.
\newblock Mocoda: Model-based counterfactual data augmentation.
\newblock In {\em Proc.of the 36th Advances in Neural Information Processing Systems (NeurIPS 2022)}, volume~35, pages 18143--18156, 2022.

\bibitem[\protect\citeauthoryear{Shi \bgroup \em et al.\egroup }{2023}]{skimo}
Lucy~Xiaoyang Shi, Joseph~J Lim, and Youngwoon Lee.
\newblock Skill-based model-based reinforcement learning.
\newblock In {\em Proc.of the 7th Conference on Robot Learning (CoRL 2023)}, pages 2262--2272. PMLR, 2023.

\bibitem[\protect\citeauthoryear{Sohn \bgroup \em et al.\egroup }{2015}]{cvae}
Kihyuk Sohn, Honglak Lee, and Xinchen Yan.
\newblock Learning structured output representation using deep conditional generative models.
\newblock In {\em Proc.of the 29th Advances in neural information processing systems (NeurIPS 2015)}, volume~28, 2015.

\bibitem[\protect\citeauthoryear{Sutton \bgroup \em et al.\egroup }{1999}]{HRL}
Richard~S Sutton, Doina Precup, and Satinder Singh.
\newblock Between mdps and semi-mdps: A framework for temporal abstraction in reinforcement learning.
\newblock {\em Artificial intelligence}, 112(1-2):181--211, 1999.

\bibitem[\protect\citeauthoryear{Wang \bgroup \em et al.\egroup }{2021}]{wang2021offline}
Jianhao Wang, Wenzhe Li, Haozhe Jiang, Guangxiang Zhu, Siyuan Li, and Chongjie Zhang.
\newblock Offline reinforcement learning with reverse model-based imagination.
\newblock In {\em Proc.of the 35th Advances in Neural Information Processing Systems (NeurIPS 2021)}, volume~34, pages 29420--29432, 2021.

\bibitem[\protect\citeauthoryear{Wang \bgroup \em et al.\egroup }{2023}]{wang2023d}
Caroline Wang, Garrett Warnell, and Peter Stone.
\newblock D-shape: Demonstration-shaped reinforcement learning via goal-conditioning.
\newblock In {\em Proc.of the 22nd International Conference on Autonomous Agents and Multiagent Systems (AAMAS 2023)}, pages 1267--1275, 2023.

\bibitem[\protect\citeauthoryear{Yang \bgroup \em et al.\egroup }{2021}]{WGCSL}
Rui Yang, Yiming Lu, Wenzhe Li, Hao Sun, Meng Fang, Yali Du, Xiu Li, Lei Han, and Chongjie Zhang.
\newblock Rethinking goal-conditioned supervised learning and its connection to offline rl.
\newblock In {\em Proc.of the 9th International Conference on Learning Representations (ICLR 2021)}, 2021.

\end{thebibliography}

\end{document}